\documentclass[12pt]{article}

\usepackage{latexsym}

\long\def\nop#1{}

\def\comment{\edef\cps{\the\parskip} \parskip=0.5cm \begingroup \tt}

\expandafter\ifx\csname shortcite\endcsname\relax

\fi

\newbox\current

\long\def\plframebox#1{
\setbox\current\vbox{#1}		

\vbox to \ht\current {\hrule\vss
\hbox to \wd\current {%
\vrule \hss\box\current\hss \vrule}
\vss\hrule }
}

\long\def\eatpar#1{%
\ifx#1\par                      
\let\nextmove=\eatpar           
\else
\let\nextmove=#1
\fi
\noexpand\nextmove
}

\def\modifymargins#1#2{
\newdimen\addtoh
\newdimen\addtow
\addtoh=#1
\addtow=#2

\advance\topmargin by -\addtoh
\multiply\addtoh by 2
\advance\textheight by \addtoh

\advance\oddsidemargin by -\addtow
\advance\evensidemargin by -\addtow
\multiply\addtow by 2
\advance\textwidth by \addtow
}

\begingroup
\catcode`\~=11
\gdef\centertilde#1{\lower #1pt\hbox{~}}
\endgroup

\newcount\currenttime
\newcount\hour
\newcount\minute

\def\printtime{%
\currenttime=\time
\hour=\currenttime
\divide\hour by 60
\minute=-\hour
\multiply\minute by 60
\advance\minute by \currenttime
\the\hour:\ifnum\minute<10 0\fi\the\minute
}

\begingroup
\makeatletter
\global\let\@@date=\@date
\gdef\@date{\@@date\ --- \printtime}
\endgroup

\def\oggi{\number\day\space 
\ifcase\month\or
Gennaio\or Febbraio\or Marzo\or Aprile\or Maggio\or Giugno\or
Luglio\or Agosto\or Settembre\or Ottobre\or Novembre\or Dicembre\fi
\space \number\year}

\newcounter{rmexample}

\def\proof{\noindent {\sl Proof.\ \ }}

\def\qed{\hfill{\boxit{}}
  \ifdim\lastskip<\medskipamount \removelastskip\penalty55\medskip\fi}
\def\qedn#1{\hfill{\boxit{}$_#1$}
  \ifdim\lastskip<\medskipamount \removelastskip\penalty55\medskip\fi}
\long\def\boxit#1{\vbox{\hrule\hbox{\vrule\kern3pt
                  \vbox{\kern3pt#1\kern3pt}\kern3pt\vrule}\hrule}}

  \def\G{{\cal G}} 
\def\I{{\cal I}}   
\def\M{{\cal M}} \def\N{{\cal N}} \def\O{{\cal O}} \def\calP{{\cal P}}

\def\ie{i.e.}
\def\eg{e.g.}

\def\aka{a.k.a.}

\def\M{{\cal M}}

\def\l{\langle}
\def\r{\rangle}

\def\C{{\rm C}}

\def\p{{\rm P}}
\def\np{{\rm NP}}

\def\S#1{\mbox{$\Sigma^p_{#1}$}}
\def\P#1{\mbox{$\Pi^p_{#1}$}}

\def\poly{{\rm poly}}

\def\pspace{{\rm PSPACE}}

\def\c{\mbox{$\leadsto$}}

\def\cp{\c{\rm P}}

\def\nuc#1{\mbox{$\parallel\!\leadsto$#1}}

\def\profont{\sf}

\def\x3c{{\profont x3c}}

\def\strips{{\profont STRIPS}}

\def\possnewtheorem#1#2{
\expandafter\ifx\csname #1\endcsname\relax
\newtheorem{#1}{#2}
\fi
}

\def\possnewtheoremthree#1[#2]#3{
\expandafter\ifx\csname #1\endcsname\relax
\newtheorem{#1}[#2]{#3}
\fi
}

\possnewtheorem{theorem}{Theorem}
\possnewtheorem{corollary}{Corollary}
\possnewtheorem{lemma}{Lemma}
\possnewtheoremthree{proposition}[theorem]{Proposition}
\possnewtheorem{definition}{Definition}
\possnewtheorem{question}{Question}
\possnewtheorem{example}{Example}
\possnewtheorem{nontheorem}{Counterexample}
\possnewtheorem{property}{Property}
\possnewtheorem{assumption}{Assumption}
\possnewtheorem{conjecture}{Conjecture}
\possnewtheorem{notation}{Notation}
\newenvironment{theorem*}[1]{{\noindent \bf Theorem~#1}\begin{it}}{\end{it}\

}

\def\plansat{{\profont PLANSAT}}

\title{On the Complexity of Case-Based Planning}
\author{Paolo Liberatore}

\begin{document}

\maketitle

\begin{abstract}

We analyze the computational complexity of problems related
to case-based planning: planning when a plan for a similar
instance is known, and planning from a library of plans. We
prove that planning from a single case has the same
complexity than generative planning (\ie, planning ``from
scratch''); using an extended definition of cases,
complexity is reduced if the domain stored in the case is
similar to the one to search plans for. Planning from a
library of cases is shown to have the same complexity. In
both cases, the complexity of planning remains, in the worst
case, \pspace-complete.

\end{abstract}
 %

\begin{verbatim}

\end{verbatim}

\section{Introduction}

Case-based reasoning \cite{kolo-93,aamo-plaz-94,wats-97} is
a problem solving methodology based on using a library of
solutions for similar problems, \ie, a library of ``cases''
with their respective solutions. Roughly speaking,
case-based planning consists into storing generated plans
and using them for finding new plans
\cite{hamm-89,berg-etal-98,spal-01}. In practice, what is
stored is not only a specific problem with a specific
solution, but also some additional information that is
considered useful to the aim of solving new problems, \eg,
information about how the plan has been derived
\cite{velo-94}, why it works \cite{kamb-90,hank-weld-95},
when it would not work \cite{ihri-kamb-97}, etc. Different
case-based planners differ on how they store cases, which
cases they retrieve when the solution of a new problem is
needed, how they adapt a solution to a new problem, whether
they use one or more cases for building a new solution, etc.
The survey papers by Bergmann et al. \cite{berg-etal-98} and
by Spalazzi \cite{spal-01} give a detailed introduction to
case-based planning.

In this paper, we study the computational complexity of
case-based planning, \ie, we characterize the complexity of
case-based planning using the theory of the \np-completeness
\cite{john-90}. What makes this analysis different from the
other results on the complexity of planning
\cite{byla-91-b,nebe-dimo-koeh-97,nebe-back-94,back-nebe-95,back-92,back-nebe-92,back-nebe-93,back-jons-95}.
is that case-based planning is not actually a {\em problem},
but rather a {\em family} of solutions for a problem. In
fact, the theory of computational complexity can only
characterize the complexity of problems, not of family of
solutions. On the other hand, case-based planning always
requires solving some specific subproblems, such as, for
example, the adaptation of a plan to a different problem. In
this paper, we study the complexity of some problems that
are related to case-based planning.

The first problem we consider is that of finding a plan
given another plan that works on a slightly different
domain. This problem formalizes the task of plan adaptation
that case-based planners have to perform. While the
formalization of this problem disregards many aspect of plan
adaptation in practice (\eg, more plans for similar cases
may be available), it nevertheless contains the main
requirement of plan adaption: building a new plan given an
old one. This problem has been already analyzed by Nebel and
Koehler \cite{nebe-koeh-93,nebe-koeh-95} under the
constraint of minimal plan change: the new plan should be as
similar to the old one as possible; plan adaptation in this
case is called {\em conservative}. There are some scenarios
in which this constraint is reasonable (\eg, we schedule the
actions of the old plan before starting planning, and not
executing them results in an additional cost); while
conservatism is sometimes obtained as a byproduct of
algorithms that work by changing an old plan \cite{kamb-90},
it is usually not a requirement. On the contrary, not
enforcing minimal change has been considered a viable
alternative for escaping the high complexity of conservative
plan adaptation \cite{au-muno-nau-02}. The absence of the
constraint of conservatism is also evident in the {\em
generative case-based} planning approach, which differs from
the {\em transformational case-based} approach outlined
above in that the new plan is generated from scratch (rather
than from the old plan), and the case is used to guide the
choices made during the search \cite{berg-etal-98}.

The main problem of analyzing a solution technique, rather
than a specific problem, is that the implemented solutions
may greatly differ to each other. While the abstract idea of
reusing old plans is part of all case-based planners, even
giving a common definition of ``an old plan'' is not easy.
This may be simply a specific plan, but can also be an
abstract plan, or a partially ordered plan, and may include
additional information \cite{kamb-90,velo-94,ihri-kamb-97}.
Moreover, the plan adaptation problem may not be tackled at
all: planning by derivational analogy
\cite{carb-86,velo-carb-93} uses the traces of the search
done for finding a plan, rather than the plan itself;
generative case-based planners \cite{berg-etal-98} build a
plan from scratch by using information from more than one
case, rather than adapting a specific plan. In the opinion
of the author of the present paper, the problem of plan
adaptation in its simpler formalization is a good starting
point for the computational analysis of plan adaptation,
even if it is not the problem that is faced in practice.

What makes giving a formalization that is at the same time
general and precise a difficult task is that plan adaptation
cannot always be separated from the other steps of
case-based planning. For example, the plan to adapt is not
chosen arbitrarily in the library of plans; it is chosen
because it is expected to be adaptable to the new situation;
moreover, more cases may be selected, leading to more than
one ``starting point'' for the search of the new plan. Being
impossible to precisely formalize the whole case-based
planning process and remain general enough, we consider the
problem of case-based planning in a very general way, in
which the only assumption that is made is that a case, or a
library of cases, is a data structure of polynomial size.

The problems that are therefore considered are: planning
from a specific known plan, and planning from a
``generalized'' case or a library of cases. The first
problem formalizes the basic plan adaptation step; the
second problem formalizes the fact that cases usually
contain additional information other than the plan itself;
the third problem is a formalization of the whole case-based
planning process.

In the analysis of these problems, we do not enforce the
constraint of minimal change. The resulting freedom in the
search for plans may simplify the algorithms
\cite{au-muno-nau-02}, but makes the computational analysis
more complicated. Indeed, the theory of \np-completeness
formalizes problems that can be expressed in a scheme like
the following one:

\begin{description}

\item[{\sc Instance:}] a planning domain;

\item[{\sc Question:}] is there a plan for the domain?

\end{description}

\noindent (for some reason not clear to the author, the
words ``instance'' and ``question'' are often written in
``small caps'' font.) The fact that this is a decision
problem (the solution can only be yes or not) is not a big
restriction, as search problems (where the solution can be a
more complex data structure) can be usually reduced to a
number of decision problems. What makes this formalization
restrictive in the setting of plan adaptation is that there
is no slot in the scheme where to place the old plan.
Indeed, the old plan is not exactly part of the instance, as
it presence does not change the definition of the problem.
As well, it is not part of the question. The plan adaptation
problem would be better formalized by a list of three parts:

\begin{description}

\item[{\sc Instance:}] a planning domain;

\item[{\sc Hint:}] a plan for a similar domain;

\item[{\sc Question:}] is there a plan for the new domain?

\end{description}

This is still a decision problem (its solution is either yes
or not). However, its definition contains a new part, the
{\em hint}, which is not necessary for answering the
question, but can be useful nevertheless. In other words,
the existence of a plan only depends on the domain, not on
the hint. The difference between the hint and the instance
makes the analysis of plan adaptation carried on in this
paper different from that of conservative plan adaptation:
in the latter problem, the plan for the similar domain is
part of the instance, as the new plan to be found depends on
the old one \cite{nebe-koeh-93}. In other words, the
instance of conservative plan adaptation contains a plan for
a domain, and the question is whether there exists a similar
plan for the new domain. Roughly speaking, an hint is
something we could disregard while looking for a solution,
while the old plan cannot in the case of conservative plan
adaptation.

The problems we analyze in this paper are the problem of
adaptation from a specific plan, from a plan plus other
information, and the problem of planning with a library of
cases. In all these cases, we are given an instance of the
planning problem and an hint, which can be a plan, a plan
plus other information, or a plan library. The technical
results of this paper is that plan adaptation of a single
specific plan can be as hard as generative planning (\ie,
planning ``from scratch'' disregarding the hint completely);
planning from a more general definition of cases may be
easier than generative planning in some cases, but is as
hard in general. The same results hold for case-based
planning, \ie, planning from a library of cases.

 %

\section{Preliminaries}

The planning problems analyzed in this paper are formalized
using the propositional \strips\  formalism. A \strips\
instance (or a \strips\  domain) is a 4-tuple $\l {\cal P},
\O, \I, \G \r$, where ${\cal P}$ is the set of conditions,
$\O$ is the set of operators, $\I$ is the initial state, and
$\G$ is the goal. The {\em conditions} are facts that can be
true or false in the world of interest. A {\em state} $S$ is
a set of conditions, and represents the state of the world
in a certain time point. The conditions in $S$ are those
representing facts that are true in the world, while those
not in $S$ represent facts currently false.

The {\em initial state} is a state, thus a set of
conditions. The {\em goal} is specified by giving a set of
conditions that should be achieved, and another set
specifying which conditions should not be made true. Thus, a
goal $\G$ is a pair $\l \M,\N \r$, where $\M$ is the set of
conditions that should be made true and $\N$ is the set of
conditions that should be made false.

The {\em operators} are actions that can be performed to
achieve the goal. Each operator is a 4-tuple $\l \phi, \eta,
\alpha, \beta \r$, where $\phi$, $\eta$, $\alpha$, and
$\beta$ are sets of conditions. When executed, such an
operator makes the conditions in $\alpha$ true, and those in
$\beta$ false, but only if the conditions in $\phi$ are
currently true and those in $\eta$ are currently false. The
conditions in $\phi$ and $\eta$ are called the positive and
negative {\em preconditions} of the operator. The conditions
in $\alpha$ and $\beta$ are called the positive and negative
{\em effects} or {\em postconditions} of the operator.

Given an instance of a \strips\  planning domain $\l {\cal
P}, \O, \I, \G \r$, we define a plan for it as a sequence of
operators that, when executed in sequence from the initial
state, lead to a state where all the conditions in $\M$ are
true and all those in $\N$ are false. More details about the
definition of \strips\  can be found in \cite{fike-nils-71}
and \cite{byla-91-b}.

\

A {\em planning case} is a pair $\l D_o,P_o \r$, where $D_o$
is a planning instance and $P_o$ is a plan for it. Plan
adaptation is formalized as follows: we are given a plan
case and a specific domain $D$ that contains the same
conditions and actions of the domain $D_o$. The problem is
that of finding a plan for $D$. This is a simplification of
the plan adaptation step of case-based planners, but
contains all main components: the case with a solution and a
new instance to be solved. Somehow, we are studying this
problem in isolation from the general case-based planning
process, as for example we disregard the fact that the case
is typically chosen in such a way $P_0$ is expected to be
useful for finding a plan for $D$. We then consider the
problem of plan adaptation in the assumption that the case
is not only a pair $\l D_o,P_o \r$ but may also contain
other information; we abstract over the kind of information
that is stored, so that our results are independent on
whether we use derivational traces, abstractions, previous
failures, etc. We prove that planning adaptation is
feasible, in this case, provided that the domain of the case
and the current domain are similar.

The second problem we consider is whether a library of plans
can improve the efficiency of finding a plan for a new
domain. A simple (and somehow simplistic) definition of a
plan library is a set of cases $\{ \l D_i,P_i \r ~|~ 1 \leq
i \leq m \}$. We consider this definition of plan library as
a special case of some importance, but the result we prove
holds for any kind of plan library. This generalization is
necessary because a case library usually contains much more
information than just a set of planning instances with their
relative solutions (\eg, plans may be abstract, they may be
partially ordered, the library may contains indexes using
for example description logics \cite{koeh-96}, the cases
themselves may be abstract and stored hierarchically, etc.)
By assuming that the library of cases is an arbitrary data
structure, our results carry on to all specific classes of
plan libraries. We remark that the complexity-theoretic
results for this generalized definition of case library are
more related to the generative case-based planning than to
the transformational approach, in that the plan library is
only assumed to be used when it is useful in the process of
searching for a new plan, and is not necessarily used to
provide a starting point of the search process.

 %

\section{Results}

The problem of deciding whether there exists a plan for a
\strips\  instance is denoted as \plansat, and is known to
be \pspace-complete \cite{byla-91-b,byla-94}. If a planning
case $\l D_o,P_o \r$ is also known, the problem cannot
become more difficult, as we can simply disregard the case
and find the plan using $D$ only. Essential to this trivial
result is that we do not enforce adaptation to be
conservative, as motivated in the Introduction. The problem
of plan adaptation is therefore in \pspace. While a problem
cannot be made harder by the presence of an hint, it may
become easier. The following theorem shows that this is not
the case for plan adaptation.

\begin{theorem}
\label{simple-fixed-plans}

Deciding whether there exists a plan for a \strips\ instance
$D$, given a case $\l D_o,P_o \r$, is \pspace-complete, even
if $D_o$ and $D$ only differ from one condition of the
initial state.

\end{theorem}

\proof The problem is in \pspace, as we can find the a plan
for $D$ disregarding the case $\l D_o,P_o \r$.

We prove that the problem of plan adaptation is \pspace-hard
by showing a reduction from \plansat. In other words, we
show that, given an instance of \plansat\  $\l
\calP,\O,\I,\G \r$, there exists $\l D_o,P_o \r$ and $D$
such that $P_o$ is a plan for $D_o$, the domains $D_o$ and
$D$ only differ for one condition of the initial state, and
$D$ has a plan if and only if $\l \calP,\O,\I,\G \r$ has a
plan.

The domains $D_o$ and $D$ are obtained from $\l
\calP,\O,\I,\G \r$ by adding a single condition $a$, a
single action $e = \l \{ a \} , \emptyset , \M , \N \r$,
where $\G = \l \M,\N \r$ is the goal of the original
instance, and possibly modifying the initial state:

\begin{eqnarray*}
D_o &=& \l \calP \cup \{a\}, \O \cup \{e\}, \I \cup \{a\}, \G \r \\
D &=& \l \calP \cup \{a\}, \O \cup \{e\}, \I, \G \r
\end{eqnarray*}

The plan $P_o$ that is part of the case is $\l e \r$. It can
be easily verified that $P_o$ is a plan for $D_o$ (its
preconditions are verified in the initial state of $D_o$,
and its consequences are exactly the goal.) As a result, $\l
D_o,P_o \r$ and $D$ constitute a valid plan adaptation
instance.

A case-based planner, while looking for a plan for $D$,
would first check whether $P_o$ works in $D$ and then try to
adapt it if it does not. In this case, $P_o$ is not
executable at all in $D$, as the precondition $a$ of $e$ is
not true in the initial state of $D$. Moreover, the plan
$P_o$ is completely useless for solving $D$, as it requires
$a$ to be true, while $a$ is not not initially true in $D$
and cannot even be made true because no action in $\O \cup
\{e\}$ makes it true. As a result, $a$ can be removed
altogether from $D$, along with $e$ that is never
executable. After this change, $D$ becomes $\l
\calP,\O,\I,\G \r$, which is exactly the original instance.
We can therefore conclude that $D$ has a plan if and only if
the original instance has a plan, thus proving the
\pspace-hardness of the plan adaptation problem.~\qed

This theorem proves that checking the existence of a case
$\l D_o,P_o \r$ does not simplify the check of existence of
a plan for a domain $D$ that is similar to $D_o$. Namely,
the problem remains \pspace-complete. We remark that the
hardness part of this theorem holds {\em even} if the
instances are similar, not only in this case. This result is
therefore relevant even in those settings in which the
choice of the case to reuse is not driven by the similarity
of the domain to the one to be solved \cite{kamb-90b}.

Analyzing the proof, we observe that the reduction used for
proving the hardness part can be described as: ``given a
planning instance $\l \calP,\O,\I,\G \r$, there exists a
case $\l D_o,P_o \r$ and a domain $D$ such that...''. The
important part is that we have chosen the specific plan
$P_o$ that is part of the case. In a way, this theorem
proves that {\em there exists} a plan for $D_o$ that does
not simplify the plan existence problem for $D$, \ie, some
plans do not help.

A natural question is now: given that some plans do not
help, may it be that some other plans do? In other words, it
may be that a sensible choice of a plan $P_o$ for $D_o$ is
useful for solving the problem for $D$. We give a negative
answer to this question.

\begin{theorem}\label{simple-any-plan}

The problem of checking the existence of a plan for $D$,
given a case $\l D_o,P_o \r$, where $P_o$ is any plan of
$D_o$, is \pspace-complete even if $D$ and $D_o$ only differ
from one condition of the initial state.

\end{theorem}

\proof The proof is based on the idea of reducing the
\plansat\  problem to the plan adaptation problem in which
$P_o$ is the only plan of $D_o$. By proving that $P_o$ is
not useful for solving \plansat\  for $D$, we prove that no
plan is, in general, useful.

The proof of Theorem~\ref{simple-fixed-plans} has the only
problem that $P_o=\l e \r$ may not be the only plan of
$D_o$. We therefore modify the domain $D_o$ to make it so.
Given a \plansat\  instance $\l \calP,\O,\I,\G \r$, the
domains $D$ and $D_o$ are still based on the set of
conditions $\calP \cup \{a\}$, where $a$ is a new condition.
The operators are $\O' \cup \{e\}$, where:

\begin{eqnarray*}
\O' &=&
\{
\l \alpha,\beta \cup \{a\},\gamma,\delta \r 
~|~
\l \alpha,\beta,\gamma,\delta \r \in \O
\}
\\
e &=& \l \{ a \} , \emptyset , \M , \N \r
\end{eqnarray*}

In words, we add $a$ as a negative precondition of each
operator in $\O$. The new action $e$ is defined as in the
proof of the previous theorem. The domains $D$ and $D_o$ are
defined as:

\begin{eqnarray*}
D_o &=& \l \calP \cup \{a\}, \O', \I \cup \{a\}, \G \r \\
D &=& \l \calP \cup \{a\}, \O', \I, \G \r
\end{eqnarray*}

The plan $P_o$ is $\l e \r$. This time, $P_o$ is the only
plan for $D_o$, as all other actions have $a$ as a negative
preconditions, while $a$ is initially true and no action
makes it false. On the other hand, the action $e$ is not
executable in $D$, as $a$ is initially false and no action
makes it true. As a result, we can remove $a$ and $e$ from
$D$, and this makes it identical to the original instance.
As a result, $D$ as a plan if and only if $\l \calP,\O,\I,\G
\r$ has a plan. This proves that the problem of plan
adaptation remains \pspace-hard regardless of how the plan
in the case $\l D_o,P_o \r$ is chosen.~\qed

This theorem proves that no plan for $D_o$ can be useful in
finding a plan for $D$, even if $D$ and $D_o$ only differ
for a single condition of the initial state. This result,
however, only holds when we assume that a case is exactly a
pair $\l D_o,P_o \r$, \ie, a domain and a plan for it. As
explained in the Introduction, case-based planners usually
record with plans other information such as derivational
traces, abstractions of the plan, information that
anticipates when the plan may fail, etc. This additional
information can simplify the plan adaptation process. Since
case-based planners greatly differ on what information is
recorded, we simply assume that the case is recorded as a
pair $\l D_o,A_o \r$, where $A_o$ is a polynomial-size data
structure. We can prove that such cases may be of help if
the new domain to be solved is similar to the one in a case.

The theorem showing this fact, as well as the following one,
requires a digression into the topic of problem
preprocessing. Problems such as plan adaptation can be
formalized as set of pairs $\l D_o,D \r$, where the question
is: given $D_o$, is there any $A_o$ such that solving the
question of plan existence of $D$ can be done in polynomial
time given $\l D_o,A_o \r$? In terms of problem
preprocessing, we are given $\l D_o,D \r$ and ask whether
preprocessing $D_o$ can result in a data structure $A_o$
that makes the \plansat\  problem on $D$ easy to solve, \ie,
polynomial-time. Problems for which preprocessing lowers the
complexity to \p\  form the class \cp, \aka\  comp-\p,
\aka\  compilable to \p. The following lemma shows that the
problem of plan adaptation is compilable to \p.

\begin{lemma}

The problem of checking the plan existence of $D$ is in \cp,
given $\l D_o,D \r$, where $D_o$ is a domain that only
differs from $D$ for a constant number of conditions in the
initial state and goal.

\end{lemma}

\proof Cadoli et al. \cite{cado-etal-02} proved that all
problems for which a polynomial number of varying parts (in
this case, the number of possible $D$) that correspond to
the same fixed part (in this case, $D_o$) is a constant,
then the problem is compilable to \p. This is actually the
case, as $D$ and $D_o$ only differ for a constant number $c$
of conditions in the initial states or goal; as a result,
for any $D_o$ there are only $|\calP|^c$ possible $D$. Since
$c$ is a constant, this function $|\calP|^c$ is polynomial
in the size of $D_o$ and $D$. As a result, the problem can
be compiled to \p, \ie, it is in \cp.~\qed

The following is an easy corollary of the above lemma.

\begin{theorem}

For every domain $D_o$ there exists a data structure $A_o$
such that $\l D_o,A_o \r$ allows for solving the problem of
plan existence of a domain $D$ that only differs from $D_o$
for a constant number of conditions in the initial state or
the goal in polynomial time.

\end{theorem}

\proof By the above lemma, the fixed part $D_o$ of the
problem $\l D_o,D \r$ can be preprocessed in such a way the
result of this phase is polynomial in size and allows for
solving the problem of plan existence for $D$ in polynomial
time, if $D$ and $D_o$ only differ for a constant number of
conditions of the initial state and goal. As a result, for
any given $D_o$ there is a polynomial-size data structure
$A_o$ such that $\l D_o,A_o \r$ makes polynomial the problem
\plansat\  for every $D$ that only differ to $D_o$ only for
a constant number of conditions of the initial state or
goal.~\qed

We remark that this result is theoretical, in that it
abstracts over the possible ``extended representations'' of
cases. How this result apply to the various specific
representations of cases is an open question.

The question of whether an extended representation of cases,
or a whole library of cases, is of help in finding a plan
for a new domain that can differ from the ones that have
already been analyzed for a non-constant number of
conditions can be given a negative answer by showing that
the problem of solving \plansat\  on $\l D_o,D \r$ is not
compilable to \p. We indeed prove that this problem is
\nuc\pspace-complete. This is proved by means of the
following lemma.

\begin{lemma}
\label{one-fixed}

For every operator $o$ over $\calP$, the \strips\  instance
$\l \calP,\O,\I,\G \r$ has a plan if and only if the
instance $\l \calP \cup \{y\},\O \cup \{o\}, \I',G \r$ has a
plan, where $y$ is a new condition not in $\calP$, the
operator $o'$ is $o$ with the addition of $y$ as a positive
precondition, and $\I'$ is $\I$ or $\I \cup \{y\}$ depending
on whether $o \in \O$ or not.

\end{lemma}

\proof Since $y$ is a new condition not in $\calP$, and is
only mentioned as a precondition of $o'$, no operator change
its value. As a result, it is true if and only if it is true
in the initial state. Moreover, $y$ is positive in the
initial state if and only if $o \in \O$. As a result, $o'$
is equivalent to $o$ if $o \in O$, and cannot be executed if
$o \not\in \O$. Since the addition of $y$ and $o'$ and the
possible removal of $o$ are the only changes from the first
instance to the second, the property of plan existence does
not change.~\qed

This lemma looks like a trivial property, but in fact it
says something interesting about the complexity of the
\plansat\  problem: regardless of whether $o \in \O$, we can
let $o'$ to be in $\O$, and use a variable $y$ to encode
whether $o$ in $\O$ or not. In other words, we are making an
element of the set of operators fixed, as $o'$ is added to
the set of operators regardless of whether $o \in \O$. Since
the original problem can be reduced to this new one in which
the set of operators has a fixed part, the complexity of the
new problem is at least as high as that of the original
problem. By iterating this procedure, we can make the set of
operators completely fixed. As a result, the complexity of
the problem remains the same even if the set of operators is
fixed.

The formal proof uses a sufficient condition for proving
that a problem is non compilable to \p\  called
representative equivalence.

\begin{lemma}

The problem of determining a plan for $D$, given $\l D_o,D
\r$, where $D_o$ and $D$ contain the same conditions and
operators, is \nuc\pspace-complete.

\end{lemma}

\proof Membership follows from the fact that \plansat\  is
in \pspace, and that any problem in \pspace\  is also in
\nuc\pspace.

Hardness: we show a reduction from the \plansat\  problem
when operators are restricted to have two preconditions and
two postconditions. The problem of plan existence has been
proved \pspace-hard even under this restriction
\cite{byla-94}.

In order for using the condition of representative
equivalence \cite{libe-01}, we need first to show three
functions for the \plansat\  problem: a classification,
representative, and extension function. We make the
following choice:

\begin{eqnarray*}
Class(\l \calP,\O,\I,\G \r) &=& |\calP| \\
Repr(n) &=&
\l \{x_1,\ldots,x_n\}, \emptyset, \emptyset, \l \emptyset,\emptyset \r \r \\
Exte(\l \calP,\O,\I,\G \r, m) &=&
\l \calP \cup \{x_{|\calP|+1},\ldots,x_m\},\O,\I,\G \r
\end{eqnarray*}

In words, the class of an instance is the number of its
conditions. The representative of the class $n$ is the
instance that has $n$ conditions, no operators, and empty
initial state and goal (this instance has the empty  plan
$\l~\r$ as the initial state satisfies the goal, but this is
irrelevant.) Extending an instance is obtained by simply
adding conditions that are not then contained in any
operator nor in the initial state or goal. Technically,
these new conditions are false in the initial state and not
required to have any specific value in the goal. As a
result, $\l \calP,\O,\I,\G \r$ has a plan if and only if
$Exte(\l \calP,\O,\I,\G \r, m)$ has a plan, as required for
the extension function.

We now show a reduction that satisfies the condition of
representative equivalence. By Lemma~\ref{one-fixed}, we can
replace $\O$ with a set of operators $\O'$ which contains a
fixed operator $o'$. By iterating this reduction for {\em
each possible operator} of four conditions over $\calP$, we
end up with an instance $\l \calP \cup Y, \O_\calP, \I'', \G
\r$, where $\O_\calP$ is obtained from the set of all
possible operators of four conditions over $\calP$, and $Y$
is a set of new conditions in correspondence with the
operators of $\O_\calP$.

Here we exploit the restriction on the number of
preconditions and postconditions. The number of possible
pairs of preconditions is given by $n(n-1)$; since each
precondition can be either positive or negative, we have
exactly four combinations. As a result, the number of
possible preconditions of an operator are $4n(n-1)$,
assuming that no operator contains the same precondition
both positively and negatively. For the same reason, there
are exactly $4n(n-1)$ postconditions. As a result, there are
exactly $16n^2(n-1)^2$ possible operators over $n$
conditions. Therefore, the size of $\l \calP \cup Y,
\O_\calP, \I'', \G \r$ is polynomially larger than the size
of $\l \calP, \O, \I, \G \r$.

The instances that result from implementing this reduction
from two instances that have the same conditions are the
same. By renaming all conditions to $\{x_1,\ldots,x_n\}$,
the results of the reduction is the same if the two
instances have the same number of conditions. As a result,
this reduction satisfies the condition of representative
equivalence, which is sufficient to show that the problem
that is reduced to (\plansat) is \C-hard for any class \C\
for which the problem that is reduced from (\plansat\
again). As a result, the \plansat\  problem is
\nuc\pspace-hard.~\qed

This proof, based on Lemma~\ref{one-fixed}, has an intuitive
construction: we can progressively make $\O$ fixed while
maintaining the property of existence of plans. As a result,
the \plansat\  problem can be reduced to the \plansat\
problem where $\O$ is fixed, \ie, when $\O$ can be
preprocessed. Since making the set of conditions $\calP$
fixed is only a matter of condition renaming, this reduction
shows that the complexity of the problem does not change
even if $\calP$ and $\O$ are fixed, \ie, the fixed part of
the instance is actually fixed. This method can be used also
for other problems. However, giving a general formulation is
made difficult by the fact that ``making a part fixed''
depend on the specific problem under analysis.

The result of \nuc\pspace-hardness of \plansat\  implies
that the knowledge of a case does not reduce the complexity
of plan existence, even if the case can contain arbitrary
data besides the plan.

\begin{theorem}

The problem of checking the plan existence for $D$ does not
become polynomial even if a case $\l D_o,A_o \r$ is known,
where $D_o$ has the same conditions and operators of $D$,
and $A_o$ is an arbitrary polynomial-size data structure
depending only on $D_o$, unless the polynomial hierarchy
collapses.

\end{theorem}

\proof Assume, on the contrary, that for every $D_o$ there
exists a polynomial-sized data structure $A_o$ (\ie, the
``extended plan'') that makes solving \plansat\  on $D$ a
polynomial task. If this were the case, it would be possible
to preprocess $D_o$ obtaining $A_o$, as the preprocessing
phase is not constrained in any way but that its result must
be of polynomial size. Since $\l D_o,A_o \r$ allows for
solving \plansat\  on $D$ in polynomial time, we have that
the problem of plan existence on $\l D_o,D \r$ is in \cp.
Since the same problem is also \nuc\pspace-hard, it follows
that $\nuc\pspace \subseteq \cp$, which implies that
$\pspace/\poly \subseteq \p/\poly$ thanks to a result by
Cadoli et al. \cite{cado-etal-02}, which in turns implies
that $\S{2} \cap \P{2} = \pspace$ thanks to a result by Karp
and Lipton \cite{karp-lipt-80}. In other words, the
polynomial hierarchy collapses to its second level.~\qed

This result can also be extended to the case in which a
whole ``library of cases'' $\{\l D_i,P_i ~|~ 1 \leq i \leq m
\}$ is given, provided that all domains $D_i$ have the same
conditions and operators of the current domain $D$, and the
whole library is of polynomial size.

\begin{theorem}

The problem of checking the existence of plans for $D$
cannot be solved in polynomial time even if a library of
plans $\{\l D_i,P_i ~|~ 1 \leq i \leq m \}$ of polynomial
size is given, where all $D_i$ have the same conditions and
operators of $D$, unless the polynomial hierarchy collapses.

\end{theorem}

\proof This is only a consequence of the above theorem:
assume that $A_o = \{\l D_i,P_i ~|~ 1 \leq i \leq m \}$;
since $\l D_o,A_o \r$ does not make \plansat\  on $D$
polynomial-time, then $\{\l D_i,P_i ~|~ 1 \leq i \leq m \}$
does not make it polynomial-time either.~\qed

As it is clear, {\em some} library of plans simplify the
problem: for example, if the library contains a case $\l
D_i,P_i \r$ where $D=D_i$, a plan for $D$ is simply $P_i$.
The theorem indeed proves that such a simplification is not
possible in general. As it is also clear from the proof,
using an extended definition of cases, in which not only
plans are recorded, is not useful either. This result
requires some discussion, which will be given in the next
section.

 %

\section{Conclusions}

In this paper, we have proved two kinds of results: first,
adapting a specific plan to a new domain is as hard as
planning from scratch; second, a case or a library of cases
composed of domains, plans, and additional information, can
simplify the problem of planning in some cases, but remains
hard in general. These results are interesting because they
are {\em formally}, and not only {\em empirically}, proved.

The first kind of results are somehow not surprising, as it
has already been observed that even changing a single
condition of the initial state or goal may make a plan
completely useless. Our results are simply formal proofs of
this observation.

The second kind of results are, in the author's opinion,
more interesting. The fact that ``extended'' cases sometimes
lower the complexity of finding plans formally validates the
trend in case-based planning of storing complex information
in cases, rather than simple plans. Nevertheless, this
formal result only holds in a very general settings, in
which no assumption, besides polynomiality of space, is made
about what is stored in the case. How this result extends to
specific form of information used in case-based planners,
such as abstract plans or derivational traces, is an open
question.

The negative results about the complexity of planning from
an extended plan or a library of plans are the most
interesting ones of this paper. First, they formally prove
that case-based planning can be as hard as generative
planning; while this phenomena has been empirically observed
\cite{berg-etal-98}, it was not yet proved that it is
intrinsic to the problem and not related to the specific
implementations. Second, compared with the result on ``small
changes'', it proves that there exists a trade-off of
efficiency and size of library in case-based planning.
Again, what is interesting in this result is that it has
formally proved, not only empirically observed.

Let us compare the results of this paper with similar work
in the literature. The computational complexity of planning
has been deeply investigated by several authors, \eg,
Bylander \cite{byla-91-b,byla-94}, Nebel
\cite{nebe-dimo-koeh-97,nebe-back-94,back-nebe-95}, and
B\"ackstr\"om
\cite{back-92,back-nebe-92,back-nebe-93,back-jons-95}. All
these works are on planning from first principles, \ie, are
about planning given only a planning domain. More related to
the present work are that by Nebel and Koehler
\cite{nebe-koeh-93,nebe-koeh-95}, who analyzed the
complexity of conservative plan adaptation. Their complexity
results are about planning when the constraint to be similar
to another known plan is enforced. While there are some
scenarios where this constraint is important, it seems not
to be enforced often in case-based planning. Some work on
plan adaptation has also been done by Bylander
\cite{byla-96}, who has shown that, probabilistically, plan
modification is simpler than plan generation provided that
the domain of the case is similar to the one to be solved.

As it has already been noticed in the introduction, the
problem of plan adaptation and that of planning with a
library of cases is of a different kind of most problems
that are studied in computational complexity, in that the
data we are given include an ``hint'' that can very well be
neglected if necessary. Such problem format include other
problems, such as case-based reasoning in general. The
complexity of other forms of case-based reasoning is an
interesting problem which is however out of the scope of
this paper.

 %

\bibliographystyle{plain}

\end{document}